\documentclass[fleqn,10pt]{wlscirep}
\usepackage[utf8]{inputenc}
\usepackage[T1]{fontenc}
\title{Modeling Visual Memorability Assessment with Autoencoders Reveals Characteristics of Memorable Images}

\author[1,2,$\dagger$]{Elham Bagheri}
\author[1,2,*]{Yalda Mohsenzadeh}

\affil[1]{Vector Institute for Artificial Intelligence, W1140-108 College Street, Schwartz Reisman Innovation Campus, Toronto, M5G 0C6, ON, Canada}
\affil[2]{Department of Computer Science, Western University, Middlesex College, Richmond St, London, N6A 5B7, ON, Canada}

\affil[*]{ymohsenz@uwo.ca}
\affil[$\dagger$]{elham.bagheri@vectorinstitute.ai}

\keywords{image memorability, visual perception, autoencoders, latent representation, interpretability analysis, distinctiveness}

\begin{abstract} % move figures and tables to the end. Add page and line numbers
Image memorability refers to the phenomenon where certain images are more likely to be remembered than others. It is a quantifiable and intrinsic image attribute, defined as the likelihood of an image being remembered upon a single exposure. Despite advances in understanding human visual perception and memory, it is unclear what features contribute to an image's memorability. To address this question, we propose a deep learning-based computational modeling approach. 
We employ an autoencoder-based approach built on VGG16 convolutional neural networks (CNNs) to learn latent representations of images. The model is trained in a single-epoch setting, mirroring human memory experiments that assess recall after a single exposure. We examine the relationship between autoencoder reconstruction error and memorability, analyze the distinctiveness of latent space representations, and develop a multi-layer perceptron (MLP) model for memorability prediction. Additionally, we perform interpretability analysis using Integrated Gradients (IG) to identify the key visual characteristics that contribute to memorability.
Our results demonstrate a significant correlation between the images' memorability score and the autoencoder's reconstruction error, as well as the robust predictive performance of its latent representations. Distinctiveness in these representations correlated significantly with memorability. Additionally, certain visual characteristics were identified as features contributing to image memorability in our model.
These findings suggest that autoencoder-based representations capture fundamental aspects of image memorability, providing new insights into the computational modeling of human visual memory. 
\end{abstract}
\begin{document}

\flushbottom
\maketitle
\thispagestyle{empty}

\section*{Introduction}\label{introduction}
Understanding why certain images are more memorable than others is a key question in both cognitive science and computer vision. In a foundational study by Isola et al., over 2,000 scene images were presented to participants in a repeat-detection task to identify recurring images, defining memorability as the likelihood of an image being remembered after a single exposure. This study revealed significant variation in memorability scores among images, with some being more memorable than others. This variation was consistent across participants, as shown by Spearman’s rank correlation analysis (split-half rank correlation: r = 0.75), indicating intrinsic properties of images that make them memorable to a broad audience \cite{isola2013makes}. This consistency has been confirmed in several subsequent studies. Goetschalckx et al. confirmed that memorability is stable across different memory tasks and intervals \cite{goetschalckx2019incidental}. Bainbridge et al. demonstrated the consistency of memorability across individuals and viewing conditions, particularly with faces \cite{BAINBRIDGE20191}. Bylinskii et al. further validated that memorability is resilient to contextual changes \cite{bylinskii2015intrinsic}, while Khosla et al.confirmed its robustness across large datasets and participants \cite{khosla2015understanding}. More recently, Almog et al. showed that memorability remains consistent across individuals of varying ages \cite{almog2023memoir}. These findings collectively establish memorability as an intrinsic, measurable characteristic of images, defined by the behavioral outcomes of memory.  Despite its robustness as a psychological phenomenon, pinpointing precise attributes that contribute to an image's memorability is complex \cite{bylinskii2015intrinsic,rust2020understanding,bylinskii2022memorability}.

Advancements in artificial intelligence (AI), particularly in deep learning, have provided strong tools not only to predict memorability \cite{isola2013makes,khosla2015understanding,praveen2021resmem,needell2022embracing,hagen2023image,younesi2024predicting} but also to modify it \cite{younesi2022controlling,goetschalckx2019ganalyze}. Notably, Khosla et al. introduced MemNet, a CNN-based model that predicts image memorability by analyzing regions of images and generating memorability heatmaps \cite{khosla2015understanding}. Further advancements in the field include the work of Praveen et al., who proposed ResMem-Net, a novel deep learning architecture combining Long Short-Term Memory (LSTM) networks with CNNs. This architecture leverages information from the CNN’s hidden layers to compute an image’s memorability score, aiming to provide a more accurate estimation of memorability \cite{praveen2021resmem}. Similarly, Needell and Bainbridge evaluated several alternative deep learning models, including residual neural networks, to enhance memorability estimation \cite{needell2022embracing}. Another notable advancement is ViTMem, a model presented by Hagen and Espeseth based on Vision Transformer (ViT) architectures. ViTMem was found to be particularly sensitive to the semantic content that drives memorability in images, marking a significant step forward in the precision of memorability prediction models \cite{hagen2023image}. In the domain of face memorability, Younesi and Mohsenzadeh proposed several state-of-the-art deep neural network architectures specialized in predicting the memorability of face photographs \cite{younesi2024predicting}. The authors further developed a method to modify and control the memorability of face images using the latent space of StyleGAN \cite{younesi2022controlling}. Their work demonstrated the effectiveness of using generative models to control and manipulate memorability in a targeted manner.

Lin et al. developed a sparse coding model to analyze how compressed image representations relate to memory encoding, focusing on the reconstruction error of these representations. The model, based on ResNet-50 embeddings, explored general memory mechanisms, including how well sparse representations align with memorability scores. Although their approach provides insights into how perceptual processing interfaces with memory, it was not exclusively focused on visual memorability but rather on understanding broader memory processes in relation to perception, guided by the level-of-processing theory \cite{lin2023seeing}.

%While many models have been developed to predict memorability scores from images, the goal of this study is not to propose a new predictive framework. Instead, we leverage an autoencoder to analyze what makes images memorable by examining reconstruction errors, latent space distinctiveness, and feature representations.

In this work, we develop a computational framework to simulate subjective experiences in the visual memory experiment used to quantify image memorability. Unlike previous studies that focus on supervised memorability prediction, our approach emphasizes unsupervised learning through autoencoders, investigating whether image memorability can be inferred from reconstruction-based representations without relying on labeled datasets.
We employ transfer learning using a VGG16-based autoencoder, pre-trained on the large-scale ImageNet dataset and fine-tuned on the categorical MemCat dataset \cite{goetschalckx2019memcat}. To understand the relationship between image structure and memorability, we analyze image reconstruction error and latent space distinctiveness, hypothesizing that higher reconstruction errors—indicative of greater complexity or uniqueness—correlate with higher memorability.
We further assess the predictive capacity of latent space representations by training a multi-layer perceptron to estimate memorability scores. We hypothesize that these learned representations encode necessary visual information to distinguish between memorable and less memorable images. To identify the specific features contributing to memorability, we employ Integrated Gradients (IG), an explainable AI technique that quantitatively and visually highlights the most influential image attributes.

While many models have been developed to predict memorability scores from images, the goal of this study is not to propose a new predictive framework. Instead, we leverage an autoencoder to analyze what makes images memorable by examining reconstruction errors, latent space distinctiveness, and feature representations.
Our findings contribute to both computational modeling of memory processes and practical applications in content creation, advertising, and education, where designing memorable visuals is of critical importance.

\section*{Methods}\label{methods}

\subsection*{Proposed approach}
Autoencoders are neural networks primarily used for representation learning. By compressing input data into a lower-dimensional latent space and subsequently reconstructing it, autoencoders effectively capture the most salient features of the input.
In this study, we use autoencoders to model the visual memorability experiment. The autoencoder's ability to highlight distinctive features that differentiate one image from another, in an unsupervised way,  makes it particularly suitable for this task. We hypothesize that an image with high reconstruction error, after the model has been trained on it, indicates the presence of unique features that are difficult to capture and reconstruct, potentially contributing to its memorability. By examining the latent space representation of images, we explore how certain image characteristics are encoded and how these characteristics correlate with human memory performance. 

\subsection*{Dataset}
\begin{figure}[h!]
    \centering
    \includegraphics[width=0.6\linewidth]{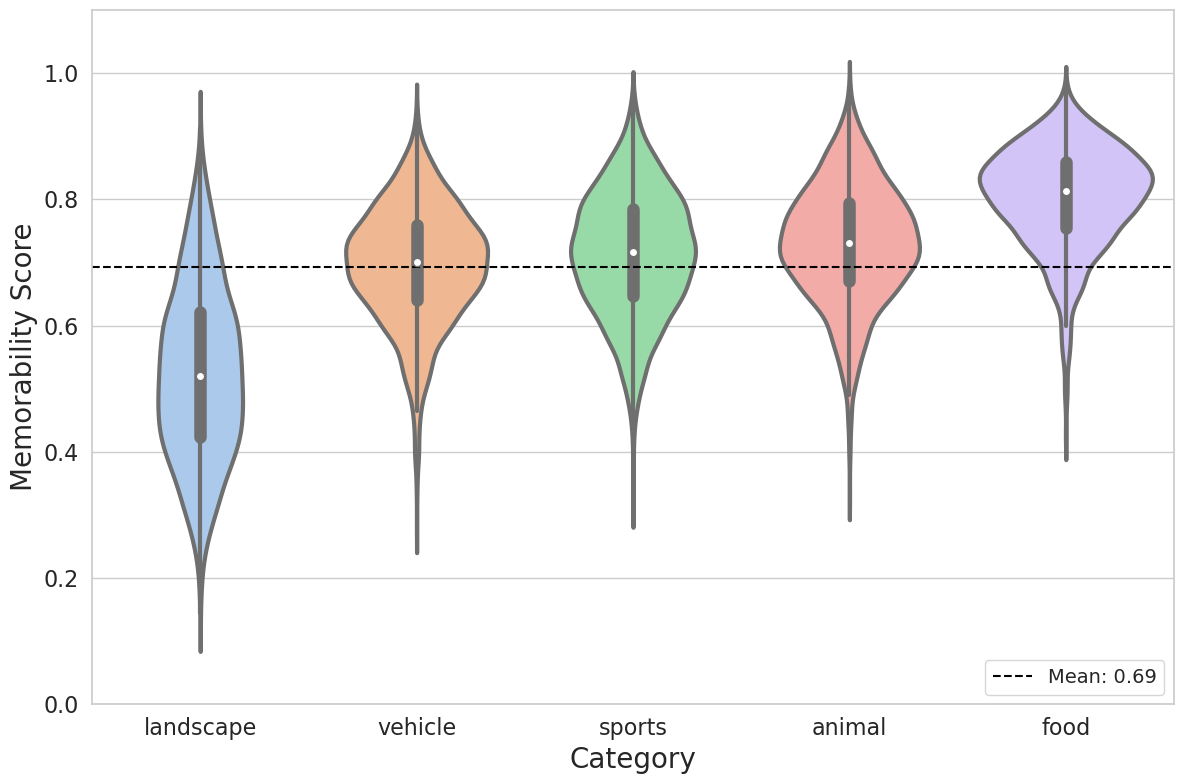}
    \caption{Distribution of false alarm-corrected memorability scores across the categories of the MemCat dataset.}
    \label{fig:memcat}
\end{figure}

The MemCat dataset, comprising 10,000 images across five categories—animals, sports, food, landscapes, and vehicles, was used for this study. Each category contains 2,000 images. The dataset's memorability scores were obtained through a repeat-detection memory game conducted on Amazon Mechanical Turk. We used MemCat's false alarm-corrected scores to account for potential biases in participant responses, ensuring a more reliable measurement of true memorability \cite{goetschalckx2019memcat,memcat}. The MemCat dataset is publicly available and can be used for academic research purposes.
Figure \ref{fig:memcat} demonstrates the distribution of memorability scores per category for the MemCat dataset.

\subsection*{Model development}

To simulate image memorability assessment, we used a CNN-based (VGG16) autoencoder pre-trained on the ImageNet dataset \cite{imagenet-autoencoder,deng2009imagenet}. VGG16 is a well-established model in computer vision \cite{simonyan2014very}, commonly used in tasks such as image classification and style transfer \cite{gatys2016image}. It utilizes deep convolutional neural networks (CNNs), which are loosely inspired by how the human visual cortex processes images. CNNs like VGG16 perform hierarchical feature extraction, starting with simple features such as edges and gradually capturing more complex patterns. The deep convolutional layers of VGG16 allow it to extract rich and efficient features from images, which aligns with our objective of understanding image memorability. The pre-trained VGG16 model simulates the accumulated visual experience of human participants.

The VGG16-based autoencoder used in this study consists of a symmetrical encoder and decoder designed to compress and reconstruct input images. The encoder mirrors the structure of the well-established VGG16 model, leveraging its hierarchical deep convolutional network design to progressively reduce the spatial dimensions of the input image while extracting increasingly complex visual features. The encoder is composed of five stages of convolutional layers with ReLU activations, with each stage increasing the number of filters while reducing the spatial resolution through max-pooling operations. Specifically, the first stage applies two 3x3 convolutional layers with 64 filters, followed by max-pooling. The second stage utilizes two 3x3 convolutional layers with 128 filters and max-pooling. The third stage incorporates three 3x3 convolutional layers with 256 filters, while the fourth and fifth stages both employ three 3x3 convolutional layers with 512 filters each, with max-pooling after every stage.

As the input image progresses through these stages, its spatial dimensions are reduced from the initial 224x224 to 7x7 in the final layer, with a depth of 512 channels. The output from this final convolutional block forms a latent space representation of size 7x7x512, which is subsequently flattened into a 25,088-dimensional vector. The encoder further processes this representation with two fully connected layers, each containing 4096 neurons and utilizing ReLU activations, with dropout applied for regularization to reduce overfitting. This encoding process compresses the input image into a compact yet information-rich latent vector that encapsulates both low-level features like edges and textures, and high-level abstractions like shapes and objects.

The decoder, which mirrors the encoder, uses transposed convolutions (also referred to as deconvolutions) to progressively upsample the compressed latent representation back to the original image dimensions. The decoder starts with three transposed convolutional layers with 512 filters, followed by another three transposed convolutional layers with 512 filters, corresponding to the deepest layers of the encoder. The subsequent stages reconstruct the feature maps using two transposed convolutional layers with 256 filters, followed by two layers with 128 filters, and finally two transposed convolutional layers with 64 filters in the output layer. A sigmoid activation function is applied at the output of the decoder to constrain the pixel values between 0 and 1, which matches the input image range.

\subsubsection*{Modeling memorability experiment}
The autoencoder was fine-tuned on the MemCat dataset for a single epoch, mimicking the human memorability experiment paradigm, where memorability is quantified by the likelihood of remembering an image after participants view images only once. The single-exposure condition, commonly employed in human memorability assessments, was thus applied to evaluate if a pre-trained autoencoder could be used to mimic human-like memorability assessment based on a single exposure to images.

All the images were resized to 224x224 pixels. Fine-tuning was performed using Various loss functions. These include optimizing both structural and perceptual similarity. The Mean Squared Error (MSE) was used to quantify pixel-wise discrepancies. Additionally, Multi-Scale Structural Similarity (MS-SSIM) \cite{wang2003multiscale}, an extension of the Structural Similarity Index (SSIM) \cite{wang2004image}, was employed to assess visual quality by analyzing structural integrity, texture, and patterns. To evaluate perceptual similarity, the Learned Perceptual Image Patch Similarity (LPIPS) metric \cite{zhang2018unreasonable} was used,  which draws upon feature representations from deep learning models such as VGG \cite{simonyan2014very}, SqueezeNet \cite{iandola2016squeezenet}, and AlexNet \cite{krizhevsky2012imagenet}. Furthermore, Style Loss (StyLoss) \cite{gatys2015neural} was employed to capture the stylistic elements of images. These diverse metrics helped to identify the best-performing models and determine which loss functions most closely mimic human visual processing relevant to memorability.

The images were fed to the model one after the other, with a batch size of 1. To ensure consistency, the order of images used to fine-tune the models was maintained across all experiments. Learning rates ranging from $10^{-7}$ to 0.1 were tested to determine the optimal settings. 

After fine-tuning, the autoencoder was tested on the same set of images, and the reconstruction error for each image was calculated as a measure of the autoencoder's ability to replicate the input after a single exposure. All experiments were repeated five times with different random seeds to ensure the robustness of the results.

% Figure~\ref{fig:schematic} demonstrates the schematic representation of our approach for modeling memorability assessment experiment by an autoencoder.

\subsection*{Memorability prediction using autoencoder latent space representations}
To investigate the potential of autoencoder latent representations for predicting image memorability, we utilized the encoder component of our fine-tuned autoencoder to extract feature embeddings from the MemCat dataset. These embeddings represent a condensed, high-level abstraction of the input images, which we hypothesized to encapsulate cognitive and perceptual factors influencing memorability.

We employed a MLP with one to three hidden layers as predictive models for memorability classification. The models were designed to refine feature mappings from the pre-trained encoder output, capturing nonlinear relationships in the latent space. 1 to 3 hidden layers of 256 to 2048 neurons were tested.

A hyperparameter search was conducted to optimize model performance, evaluating learning rates between 0.01 and 0.00001 and batch sizes of 8, 16, 32, and 64. The models were trained using the AdamW optimizer, which incorporates decoupled weight decay for better generalization. To prevent overfitting, weight decay (L2 regularization) values of 1e-5, 1e-4, and 0.01 were explored. Additionally, dropout (ranging from 0.1 to 0.5) and layer normalization were incorporated to stabilize training and mitigate covariate shift.
To ensure training efficiency and prevent overfitting, early stopping was applied with a patience of five epochs, halting training when validation loss stopped improving. 
These architectural and optimization choices ensured that the model effectively leveraged the encoder’s latent representations while maintaining robustness and avoiding overfitting.

The dataset was partitioned into training, validation, and test subsets, consisting of 5892, 931, and 3177 images, respectively, ensuring a balanced distribution with a mean memorability score of approximately 0.693.

The primary objective of this approach was to assess the capacity of the autoencoder’s latent space to encode cognitively relevant features predictive of memorability. Success in this endeavor would indicate that these latent codes effectively capture high-level visual and semantic information critical to memorability, thereby validating the encoder's ability to generate task-relevant representations and shedding light on the inherent characteristics contributing to image memorability.

% explain all other models trained and compare their AUCs
\subsection*{Explainability Analysis}
To better understand what image features contribute to memorability prediction using the latent space representations, we employed Integrated Gradients (IG), an Explainable AI (XAI) technique that attributes the prediction of a neural network to its input features by computing the gradients of the output with respect to the input at points interpolated between a given baseline and the actual input \cite{sundararajan2017axiomatic}. This method is particularly suited for our complex architecture, which includes an autoencoder followed by a feed-forward model.

The baseline input was set as a blurred mean image, computed by first averaging all training images and then applying a Gaussian blur (kernel size = 15). This ensures a smooth baseline representation that retains overall structure but removes fine details. IG attributes were computed by averaging the gradients over 50 steps between the baseline and the original image. These gradients provided a detailed map of feature contributions, helping us identify which aspects of the image were most influential in making it memorable according to the model’s assessment. The analysis was conducted using the Captum library \cite{kokhlikyan2020captum} in PyTorch.

% We also visualize the attention layers of the trained tranformer to explain ...
% \begin{figure}[t]
%     \centering
%     \includegraphics[width=1\linewidth]{schematic.png}
%     \caption{Schematic representation of the approach to model memorability assessment experiment. The autoencoder, pre-trained on ImageNet, is fine-tuned on the MemCat image memorability dataset for one epoch and batch size of one.}
%     \label{fig:schematic}
% \end{figure}

\subsection*{Feature Analysis } %for Memorability Prediction
We further explored which high-level image features contributed to the model's prediction of higher memorability. The features we considered included semantic, structural, color-based, and texture-related properties. These features were analyzed by correlating their values with both memorability scores and Integrated Gradients (IG) attributions, allowing us to assess their influence on the model's predictions.

A pre-trained ResNet-50 model was used to extract deep features, summarizing the high-level visual information present in the images \cite{he2016deep}. Specific metrics calculated included the number of distinct classes detected in each image, the total number of objects, and the average object size, computed as the mean area covered by detected objects. These measures provided insights into scene complexity and object diversity, which influence memorability by affecting perceptual load and distinctiveness.

For segmentation and object detection, we employed pre-trained DeepLabV3 and Faster R-CNN models \cite{chen2018encoder,ren2016faster} to quantify object characteristics and scene complexity. The Separation Score, derived from DeepLabV3, measured how well objects stood out from the background, capturing foreground-background contrast, which can influence visual focus and memorability. Meanwhile, the Faster R-CNN model detected objects in the image, providing metrics such as the number of objects and their average size, which further informed the role of object richness in memorability.

Color features were extracted using the CIE-LAB and HSV color spaces, which better align with human perception. We computed mean values of the L (Lightness), A (green-red), and B (blue-yellow) channels to assess dominant color tones \cite{fairchild2013color}. Saturation, representing color intensity, was measured in the HSV color space, as more vivid images tend to capture attention and enhance retention. Color diversity, computed using Shannon entropy, quantified the variation in color composition, reflecting the presence of multiple hues and measuring visual richness.

To analyze texture and structural complexity, we computed image entropy, which quantifies the randomness in pixel intensities, reflecting the information content in an image \cite{gonzalez2009digital}. Texture energy, derived from the Gray Level Co-occurrence Matrix (GLCM), captured local variations in surface details, providing insight into the richness of textures \cite{haralick1973textural}. The clutter score, calculated using edge detection via the Canny algorithm \cite{canny1986computational}, quantified the density of structural details, as cluttered images require greater cognitive processing.

Each feature's impact on image memorability was analyzed by correlating its values with average IG attributions from our trained model. This allowed us to determine which visual properties were most influential in shaping the model’s predictions and how these features aligned with human memory recall patterns.

% \sebsection*{Code availability}
% probably should do after acceptance

%% if add any supplementary: and more details on full range of hyperparameters are provided in the supplementary information. can be found as Supplementary Table S1 online./(see Supplementary Fig. S2 online)
\section*{Results}\label{results}
\subsection*{Memorable images are harder to reconstruct and more distinct} 

To evaluate whether there is a difference in the model's performance in reconstructing images with varying memorability, we calculated Spearman's rank correlation between the images' memorability scores and their reconstruction errors. This reconstruction error served as a quantitative metric to assess the model's accuracy in reproducing the input data. To ensure comparability across different loss functions, the reconstruction errors were normalized within each model to fall within the range of [0,1]. 

Our results show a significant positive correlation between the autoencoder's reconstruction error and memorability scores. The model fine-tuned using the LPIPS (SqueezeNet) loss function with a learning rate of $5E^-06$ achieved the highest correlation. This model, fine-tuned on all MemCat images, was selected for the remainder of the analysis in this paper. The Spearman’s rank correlation for this selected model was 0.452, (CI: [0.442, 0.467]), with a p-value approaching zero. 

\begin{figure}[h!]
    \centering
    \includegraphics[width=1\linewidth]{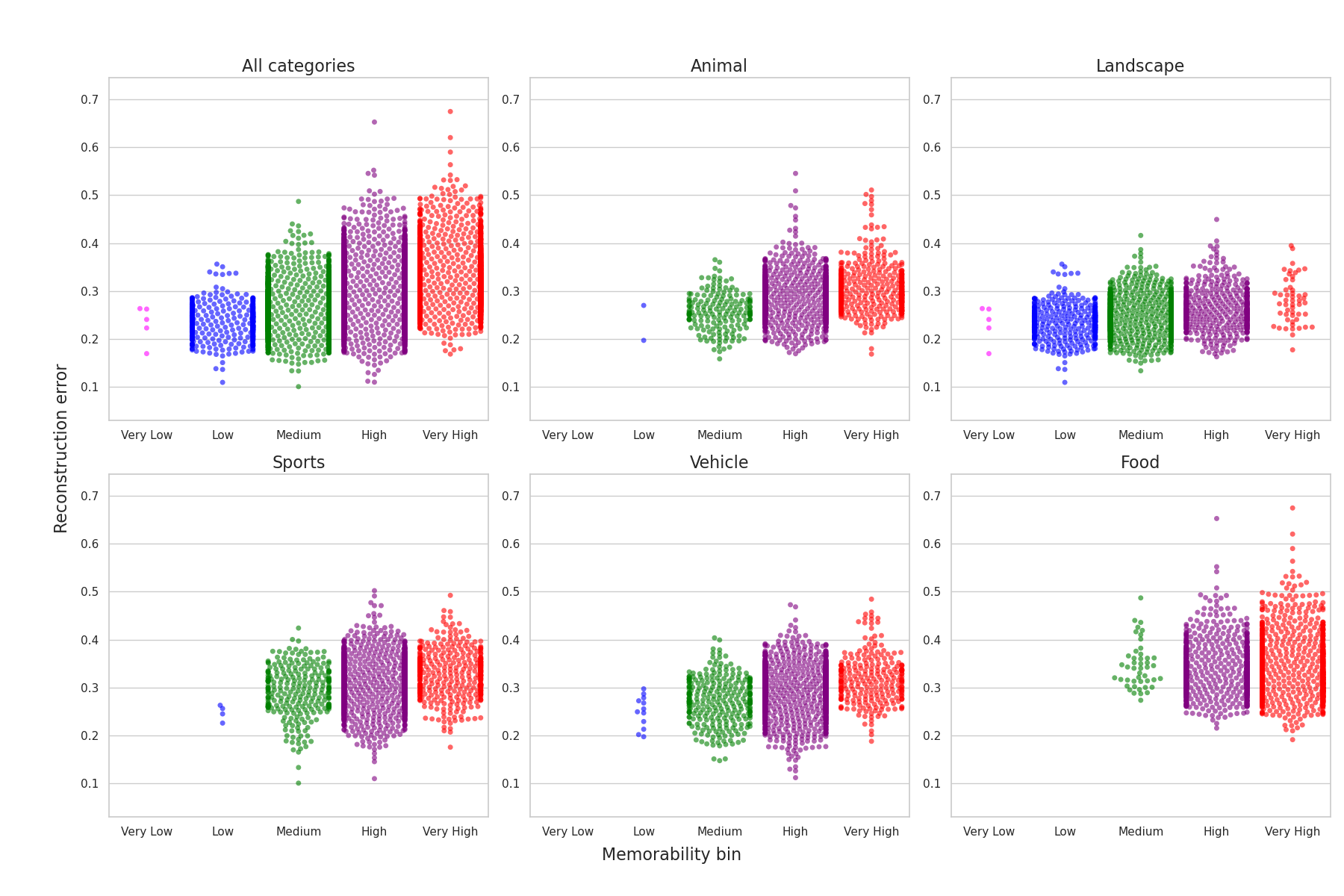}
    \caption{Distribution of reconstruction errors across varying memorability scores for every category.}
    \label{fig:error}
\end{figure}

This observation suggests that images more challenging for the autoencoder to reconstruct tend to be more memorable. Figure \ref{fig:error} illustrates the distribution of reconstruction errors across varying memorability scores with memorability scores binned at intervals of 0.2 defined as very low to very high memorable images. The figure shows the distribution of reconstruction error and shows that memorable images consistently present greater challenges for the autoencoder, this holds true for all categories, except the 'food' category.

Furthermore, we investigated the distinctiveness of images within the autoencoder's latent space, hypothesizing that more distinctive images are more likely to be memorable. Distinctiveness in memorability research is typically defined as the Euclidean distance between a target image and its nearest neighbor \cite{lin2023seeing}. However, this measure can be sensitive to the dataset's size. To address this, we introduced a normalized metric, standardizing the distances to the nearest neighbor using Z-score standardization. This provides a robust, dataset-independent metric of distinctiveness and ensures that distinctiveness remains robust across different dataset sizes, providing a stable and consistent measure of how unique each image is within its context.

\begin{figure}[h!]
    \centering
    \includegraphics[width=1\linewidth]{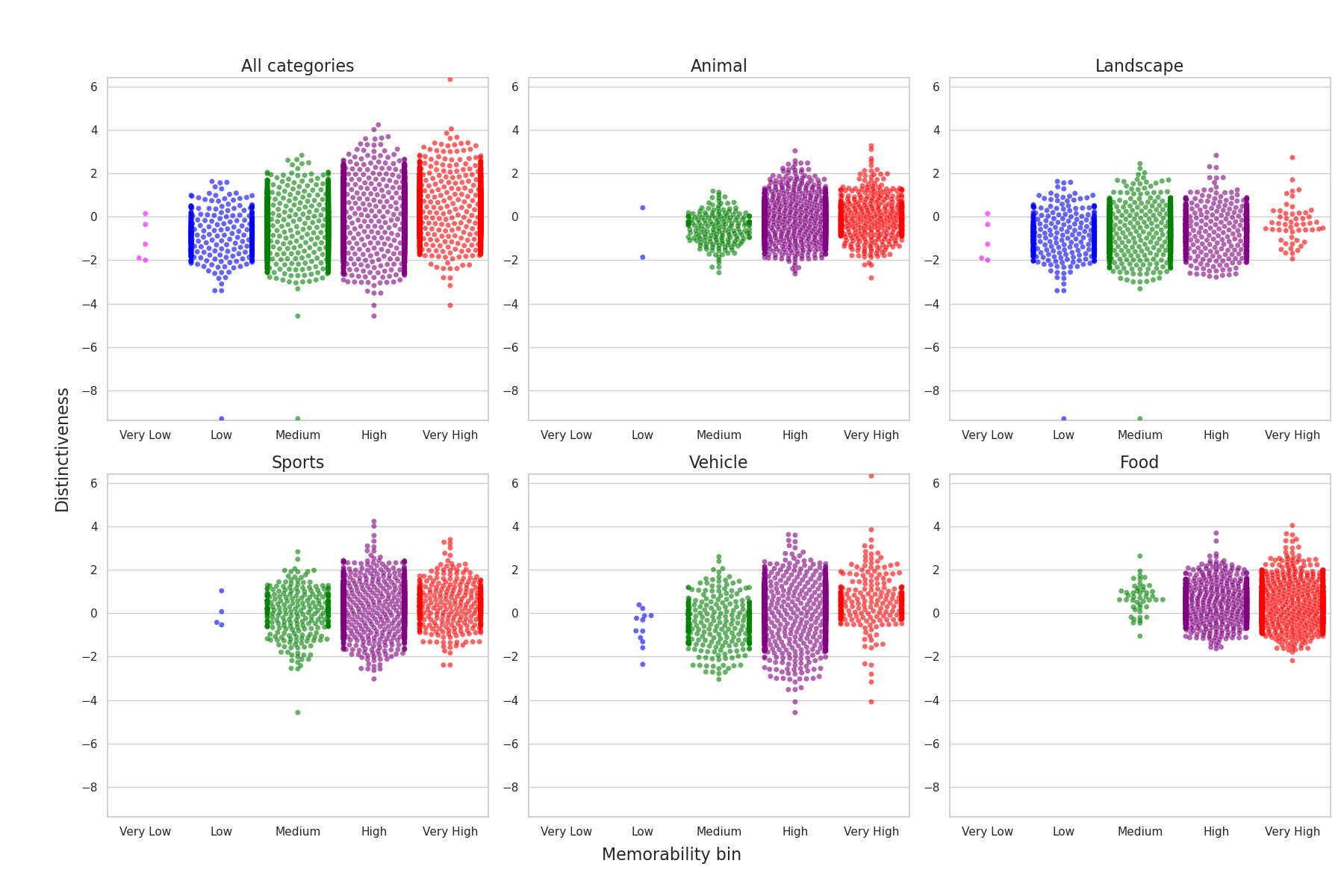}
    \caption{Distribution of the normalized distinctiveness measure across varying memorability scores for every category.}
    \label{fig:distinctiveness}
\end{figure}

Our results show that the latent space distinctiveness, as measured using the normalized metric, exhibited a positive correlation of 0.31, (CI: 0.292, 0.328), $p-value=3 \times 10^-223$, with memorability scores. This indicates that images with unique and distinct features, which are more difficult for the autoencoder to compress, are inherently more memorable. Figure \ref{fig:distinctiveness} shows the distribution of distinctiveness for binned memorability scores. This figure shows the general trend of higher distinctiveness for more memorable images for all categories except the 'food' category.

In the reconstruction error analysis (Figure \ref{fig:error}), the variability of reconstruction errors across memorability bins for food images is lower compared to other categories. While other categories show a clear increase in reconstruction error with memorability, the food category exhibits a more uniform distribution, indicating that the autoencoder's reconstruction capability is less sensitive to variations in food image memorability. This could be attributed to the visual homogeneity of food images in terms of color, texture, and composition, which may limit the model’s ability to distinguish between memorable and less memorable images.

Similarly, in the distinctiveness analysis (Figure \ref{fig:distinctiveness}), the food category shows a narrower range of distinctiveness scores across memorability bins. Although distinctiveness generally increases with memorability, this trend is subdued for food images. The limited variability in distinctiveness scores may stem from the memorability scores of food images being concentrated within a narrow range, as observed in Figure \ref{fig:memcat}. This concentration may reduce the relative distinctiveness of food images within the dataset.

Overall, the food category highlights the importance of category-specific features in determining memorability. The uniformity in visual characteristics of food images appears to weaken the correlation between memorability and both reconstruction error and distinctiveness. These findings suggest that the memorability of food images may rely on subtler cues that are not effectively captured by the current modeling approach. Future studies could explore alternative feature representations or loss functions tailored to homogeneous categories like food to enhance the model's sensitivity to variations in memorability.

\subsection*{The memorability effect diminishes with multiple exposures}
To examine how continued fine-tuning affects the observed correlation between model reconstruction error and memorability, the autoencoder was trained for up to 20 epochs. As expected, the correlation between reconstruction error and memorability decreased as the model was exposed to the images multiple times. After 20 epochs, the correlation dropped significantly to 0.22, (CI: 0.21, 0.24), with $p-value=2.8 \times 10^-114$, suggesting that repeated exposures to an image reduce its impact on memorability. 
This trend aligns with human memory studies, where repeated exposure leads to increased familiarity but reduced novelty, ultimately affecting recall performance. The model's decreasing correlation with memorability across epochs suggests that its learned representations gradually shift towards more generalizable features rather than the initial distinctive patterns that drive single-exposure memorability.

\subsection*{Models trained with perceptual loss show a stronger effect}
By comparing all fine-tuned models, we evaluated the impact of different loss functions on the model’s ability to predict memorability, as detailed in Table \ref{tab:lossfuncs}. The table presents the correlations between reconstruction errors and memorability scores for each loss function, along with confidence intervals and learning rates of the best-performing model in terms of correlations. Results show that models trained with perceptual loss functions, especially LPIPS (SqueezeNet), consistently outperformed those trained with traditional pixel-wise metrics like MSE or structural metrics like MS-SSIM.

\begin{table}[ht]
\centering
\caption{Correlation of reconstruction errors with memorability scores for models trained using different loss functions.} 
\label{tab:lossfuncs}
\begin{tabular}{@{}lccc@{}}
\toprule
Loss Function & Correlation & Confidence Interval & Learning Rate \\ \midrule
LPIPS (SqueezeNet) & 0.452 & (0.442, 0.467) & $5 \times 10^{-6}$ \\
LPIPS (VGG) & 0.434 & (0.416, 0.444) & $5 \times 10^{-4}$ \\
Style Loss & 0.371 & (0.361, 0.388) & $5 \times 10^{-6}$ \\
LPIPS (AlexNet) & 0.331 & (0.321, 0.336) & $1 \times 10^{-6}$ \\
Mean Squared Error (MSE) & 0.235 & (0.229, 0.249) & 0.1 \\
Multi-Scale SSIM & 0.152 & (0.135, 0.159) & $1 \times 10^{-7}$ \\ \bottomrule
\end{tabular}
\end{table}

\subsection*{Memorability behavior can be predicted from the autoencoder latent representation}
The best-performing model was selected based on performance on the validation set. The best prediction model was a deep neural network with two hidden layers of 1024 and 512 neurons, trained with a learning rate of \(5 \times 10^{-5}\), a batch size of 8, a dropout rate of 0.3, and layer normalization. The model was optimized using AdamW with a weight decay of 0.01 and trained with early stopping using a patience of 5 epochs. The model was trained for 23 epochs.

According to the results, our memorability prediction model demonstrated robust performance in predicting memorability from the latent representations generated by the encoder. For binary classification of memorability, on the unseen test set,the model achieved an area under the ROC curve (AUC) of 0.76 (CI: [0.74, 0.78]) and a Precision-Recall (PR) AUC of 0.74 (CI: [0.71, 0.76]). Additionally, the model's predicted scores correlated significantly with the original memorability scores with correlation = 0.56 (CI: [0.53, 0.58], \(p = 2.21 \times 10^{-258}\)), supporting the effectiveness of latent representations in capturing memorability-related features.

The binary classification of images into high and low memorability, with the threshold of the median memorability score, the model achieved an accuracy of 0.69 (CI: [0.68, 0.71]), a precision of 0.69 (CI: [0.67, 0.71]), a sensitivity of 0.70 (CI: [0.67, 0.72]), a specificity of 0.69 (CI: [0.67, 0.71]), and an F1 score of 0.69 (CI: [0.67, 0.71]).

\subsection*{Model Interpretation identifies image characteristics contributing to memorability}
\begin{figure}[h!]
    \centering 
    \includegraphics[width=1.1\linewidth]{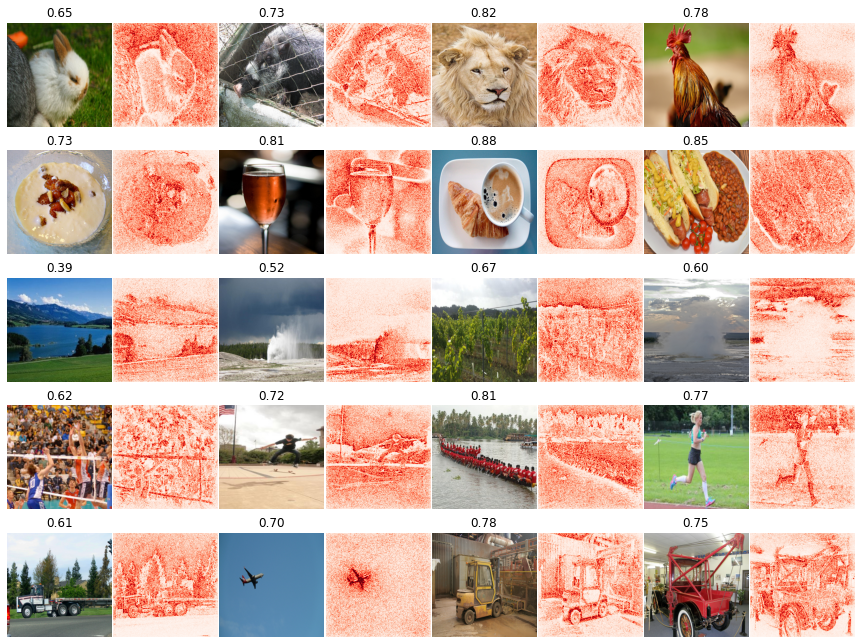}
    \caption{Integrated gradient attributes for sample images from the MemCat dataset, showing regions influencing the memorability predictor outputs. The numbers on top of the images show the Memorability score. Red areas indicate strong contributions to the model’s predictions. Each row corresponds to one of the five categories: animals, food, landscapes, sports, and vehicles, with images arranged in ascending order of memorability scores.}
    \label{fig:images_interpret}
\end{figure}

We employed the Integrated Gradients (IG) method to interpret the predictions of our memorability model, visualizing which regions in an image contributed most to its memorability score. Figure \ref{fig:images_interpret} illustrates the IG attributions for selected images from the MemCat dataset, where red regions indicate areas with higher attribution values. These regions are identified as the most influential for the model’s prediction of memorability.

In the animal category, such as images of the rabbit, lion, and rooster, the IG heatmaps emphasize facial regions, particularly the eyes and prominent features like fur patterns. These observations suggest that distinct facial features and textures play a crucial role in making animals memorable. For example, the lion’s mane and the rooster’s sharp contours further underscore the importance of high-contrast and well-defined features in driving memorability.

For food-related images, including yogurt, wine, coffee, and hot dogs, the IG attributions concentrate on vivid textures, strong contrasts, and central elements. The smooth surface of the wine glass and the detailed textures of the food toppings are highly attributed. This highlights the role of vibrant colors, intricate patterns, and well-composed central features in influencing memorability in food images.

In landscape scenes, such as mountains, geysers, and vineyards, the IG heatmaps show strong attributions for horizon lines, separation between foreground and background, and specific details like trees or water surfaces. The prominence of natural structures and high-contrast boundaries suggests that such elements contribute significantly to the memorability of landscape images.

For sports and activity images, such as basketball players, rowers, and runners, the IG attributions highlight human figures and interactions. The focus on human faces, gestures, and motion-related elements, such as the oars and leg movements, underscores the model's sensitivity to dynamic, socially relevant features. The emphasis on human-related elements demonstrates their intrinsic link to memorability.

In the vehicle category, including airplanes, trucks, and machinery, the IG maps emphasize distinct shapes and structural details, such as wheels, propellers, and edges of the vehicle. These findings suggest that clear geometric patterns and mechanical features enhance memorability in vehicle-related imagery.

Overall, the IG analysis reveals that the model consistently focuses on distinct, high-contrast regions, central objects, and unique features across all categories. Elements like bright colors, textures, and recognizable structures (e.g., faces, objects) emerge as critical contributors to memorability. Furthermore, the attribution maps highlight the role of image composition, with the arrangement of features and their saliency playing an essential role in drawing attention and enhancing recall. Notably, the distribution of attributions differs slightly between categories. While human-related and animal features often dominate their respective categories, natural landscapes and vehicles rely more on structural complexity and contrast. This variability suggests that memorability is influenced not only by individual features but also by the contextual arrangement and saliency of elements within an image.

\subsection*{Feature Importance in Predicting Memorability from Latent Representations}
The feature importance analysis in this study provides a detailed understanding of which image attributes significantly contribute to memorability. Using correlations with Integrated Gradient (IG) attributions and original memorability scores, we identified several features from the MemCat dataset that aligned well with human-perceived memorability. The features having significant correlation values with original memorability scores and IG attributes are listed in Table \ref{tab:feature_correlations}. The p-values indicate the statistical significance of these correlations. These findings provide insights into the visual attributes that influence image memorability and validate the relevance of the autoencoder-based latent representations used in the prediction model.

The results indicate that object presence and arrangement strongly affect memorability. Images containing a greater number of distinct classes, a higher number of objects, and larger objects tend to be more memorable. This suggests that increased scene complexity and diversity provide more reference points for memory encoding. Larger objects likely draw more visual attention, while distinct elements within an image may facilitate richer encoding and longer retention.

Visual complexity also plays a critical role. The clutter score, derived from edge detection, quantifies the presence of structural details in an image. The positive correlation with memorability suggests that denser, detail-rich images require greater cognitive processing, making them more likely to be retained. Image entropy, which quantifies randomness in pixel intensities, further supports this observation. Higher entropy values indicate greater visual information content, reinforcing the idea that diverse textures and non-uniform compositions enhance memorability.

Color properties contribute significantly to image retention. The strong correlation between memorability and the LAB color channels suggests that color balance across green-red (A) and blue-yellow (B) axes influences memorability. Higher values along these axes indicate distinct and contrasting color compositions, which are more likely to capture attention. Higher saturation was also positively correlated with memorability, confirming that vivid and intense colors enhance image recall. Conversely, the negative correlation with mean color LAB L (lightness) suggests that excessively bright images may reduce contrast, making objects within them less distinguishable and, consequently, less memorable.

The separation score, measuring foreground-background distinction, was positively associated with memorability. This suggests that images where the main subject stands out clearly from the background are more likely to be remembered, likely due to enhanced object recognition and reduced perceptual competition.

Texture-related features exhibited mixed effects. Texture energy, which reflects uniformity in patterns, was negatively correlated with memorability, indicating that highly uniform textures may lack distinctiveness and fail to engage visual processing. Higher levels of structural complexity, captured through clutter and edge density, were positively correlated, suggesting that diverse and complex textures create additional memory cues.

The comparison between IG attributions and original memorability scores revealed a consistent trend in feature importance. Features that strongly correlated with memorability scores, such as the number of distinct objects, object size, and separation score, also exhibited high correlations with IG attributions. This suggests that the model’s predictions align with human perception of memorability, validating the autoencoder-based latent representations as effective for memorability modeling.

While most features maintained similar correlations across both measures, some variations were observed. Saturation and image entropy exhibited stronger correlations with IG attributions than with original memorability scores, indicating that the model assigns more importance to these properties during prediction than humans may consciously recognize. Conversely, some structural complexity measures, such as texture energy, had weaker correlations with IG attributions, suggesting that although these features contribute to memorability, they are not directly emphasized by the model in prediction tasks.

Overall, these findings highlight that a combination of object properties, color composition, structural complexity, and foreground-background separation influences memorability. The alignment between IG attributions and original memorability scores suggests that the model effectively captures the same visual attributes that determine human memory retention.

\begin{table}[h]
\centering
\caption{Spearman's rank correlation of image features with integrated gradients attributes and original memorability scores, including confidence intervals and p-values.}
\label{tab:feature_correlations}
\begin{tabular}{lcccccc}
\toprule
\textbf{Feature} & \multicolumn{3}{c}{\textbf{Integrated Gradients Attributes}} & \multicolumn{3}{c}{\textbf{Memorability Scores}} \\
\cmidrule(lr){2-4} \cmidrule(lr){5-7}
& \textbf{Corr.} & \textbf{95\% CI} & \textbf{p-val.} & \textbf{Corr.} & \textbf{95\% CI} & \textbf{p-val.} \\
\midrule
Mean color LAB A & 0.585 & [0.557, 0.608] & $6.47 \times 10^{-291}$ & 0.347 & [0.314, 0.377] & $1.27 \times 10^{-90}$ \\
Distinct object categories & 0.531 & [0.506, 0.556] & $1.15 \times 10^{-230}$ & 0.440 & [0.411, 0.471] & $1.32 \times 10^{-150}$ \\
Num. objects & 0.463 & [0.437, 0.489] & $9.79 \times 10^{-169}$ & 0.350 & [0.316, 0.380] & $5.88 \times 10^{-92}$ \\
Mean color LAB B & 0.418 & [0.390, 0.450] & $2.10 \times 10^{-134}$ & 0.228 & [0.194, 0.260] & $9.32 \times 10^{-39}$ \\
Saturation & 0.413 & [0.384, 0.445] & $2.05 \times 10^{-131}$ & 0.157 & [0.122, 0.191] & $6.59 \times 10^{-19}$ \\
Average object size & 0.404 & [0.374, 0.432] & $9.29 \times 10^{-125}$ & 0.480 & [0.449, 0.510] & $3.63 \times 10^{-183}$ \\
Separation score & 0.321 & [0.287, 0.351] & $6.69 \times 10^{-77}$ & 0.273 & [0.241, 0.305] & $1.91 \times 10^{-55}$ \\
Color diversity & 0.276 & [0.244, 0.308] & $1.04 \times 10^{-56}$ & 0.115 & [0.084, 0.154] & $9.42 \times 10^{-11}$ \\
Image entropy & 0.160 & [0.128, 0.193] & $1.21 \times 10^{-19}$ & 0.102 & [0.066, 0.134] & $7.36 \times 10^{-9}$ \\
Clutter score & 0.136 & [0.102, 0.169] & $1.35 \times 10^{-14}$ & 0.124 & [0.086, 0.158] & $2.79 \times 10^{-12}$ \\
Mean color LAB L & -0.123 & [-0.156, -0.088] & $3.61 \times 10^{-12}$ & -0.058 & [-0.092, -0.022] & $9.82 \times 10^{-4}$ \\
Texture energy & -0.108 & [-0.140, -0.075] & $1.10 \times 10^{-9}$ & -0.078 & [-0.113, -0.046] & $9.57 \times 10^{-6}$ \\
\bottomrule
\end{tabular}
\end{table}

\section*{Discussion}\label{discussion}

We observed a strong positive correlation between the model reconstruction error of images and image memorability scores, supporting the hypothesis that more memorable images are inherently more challenging for the autoencoder to reconstruct. Prior works have modeled memorability using classification models or generative approaches \cite{kyle2019predicting, lin2023seeing}. In contrast, our study focuses on leveraging autoencoder-based feature representations to understand why certain images are harder to reconstruct and how this relates to memorability. Further, our results showed that the distinctiveness of images, as captured by the autoencoder’s latent space, also correlates with memorability. This strong positive correlation indicates that images with more pronounced features, represented by higher normalized distances from their nearest neighbors, are more memorable. This suggests that memorability may not be solely dependent on the presence of specific objects or textures but emerges from the structural and compositional uniqueness of an image, making it more difficult for the model to encode efficiently.

We further observed that the correlation between reconstruction error and memorability diminished as the model was exposed to the images repeatedly. This result suggests that the novelty or distinctiveness of an image, which contributes to its memorability, decreases with repeated exposure. This aligns with theories in cognitive psychology indicating that repeated exposure leads to habituation, reducing an image’s perceptual saliency over time. This finding generates a new testable hypothesis: repeated exposure reduces an image's memorability by diminishing its distinctiveness in the latent space. Future experiments can explore how varying levels of exposure affect the underlying features that contribute to memorability, providing deeper insights into the relationship between exposure, distinctiveness, and memory retention.

Our approach supports previous studies that have suggested items that stand out from their context are better remembered \cite{bylinskii2015intrinsic,attneave1959applications,eysenck2014depth,hunt2006distinctiveness,konkle2010conceptual,rawson2008does,schmidt1985encoding,standing1973learning,wiseman1974perceptual,vogt2007long,von1933wirkung}. However, unlike these earlier studies, our model enables an unsupervised, computational evaluation of memorability, moving beyond human-centric assessment methods and allowing large-scale, automated analysis of visual memory encoding.

Our findings on the correlation of reconstruction errors and image memorability are closely related to the work by Lin et al. \cite{lin2023seeing}. They introduced a sparse coding model that used reconstruction error and distinctiveness as metrics for memory formation and retention, focusing on general perceptual memory processes rather than image memorability specifically. While both studies identify reconstruction error as an important predictor of memorability, our approach differs in several ways. First, we apply autoencoders rather than sparse coding, allowing us to reconstruct images directly while analyzing their latent space representations. Second, our work focuses explicitly on image memorability rather than broader perceptual memory encoding, demonstrating that autoencoder reconstruction error can serve as a direct proxy for memorability. Finally, whereas Lin et al. focused on compressing feature embeddings extracted from an object recognition model, our model directly quantifies memorability as an intrinsic image property.

The positive correlation between the autoencoder reconstruction error and memorability score was consistently observed across the different loss functions we tested. However, the highest correlation was achieved when perceptual-based loss functions were used to train the model. This finding shows the importance of selecting appropriate loss functions when modeling human cognitive processes. This result also suggests that future modeling approaches aiming to capture human cognitive patterns should prioritize perceptual alignment rather than pixel-level fidelity, especially in domains like visual memory and perception.
% This finding has broader implications for designing machine learning systems that aim to replicate human cognitive processes, particularly in domains like visual memory and perception. 

The reasonable accuracy of the prediction model on the autoencoder’s latent representations in predicting memorability further demonstrated that the compressed representations generated by the autoencoder contain information to predict memorability scores, suggesting the strength of latent space representations as a proxy for human memory assessment. The ability of the model to capture essential features for memorability suggests that latent space representations encode vital information that correlates strongly with human memory recognition.

% This aligns with theories in neuroscience and AI that suggest human memory relies on compressive encoding of distinct, high-information features, a principle mirrored in how deep networks represent visual content.

In addition, the use of Integrated Gradients (IG) for model interpretation provided new insights into the visual features most influential in making an image memorable, addressing a key limitation in earlier works that lacked model interpretability. By directly attributing memorability predictions to specific visual features, our results provide a more concrete mapping between computational representations and human perception. 

Our findings are supported by several established studies in the field. For example, Isola et al. found that semantic content such as the presence of objects and scenes plays a significant role in predicting memorability, with certain types of content (e.g., human figures, interiors) being more memorable than others (e.g., natural scenes, backgrounds) \cite{isola2013makes}. Kholsa et al. mention that memorable images tend to include distinct objects, people, and salient actions or events, while forgettable images are often characterized by natural landscapes or scenes that lack distinguishable features \cite{khosla2015understanding}. Bylinskii et al. also noted that overly simple or uniform images tend to be less memorable \cite{bylinskii2015intrinsic}.

An exception to the observed trends was found in the food category, where the correlation between memorability and both reconstruction error and distinctiveness was lower than in other categories. Unlike objects, landscapes, or human figures, food images exhibit greater visual homogeneity in color, texture, and composition, which likely reduces their distinguishability in the model’s latent space. This suggests that food memorability may depend more on semantic, cultural, or personal associations rather than purely visual distinctiveness. Future work should explore incorporating multimodal embeddings (e.g., text-based descriptions, prior exposure modeling) to better capture non-visual memorability cues in such categories.

\textbf{Limitations and future directions:} While our findings are significant, there are certain limitations to our approach. 
First, the MemCat dataset, though diverse, may not fully capture the full range of image types and contexts that are relevant to memorability. Future studies should explore broader and more diverse datasets to validate and extend these findings. Second, the use of a CNN-based autoencoder, though effective, may not be optimal for all types of images. Brain-inspired architectures or other deep learning models should be explored in future research to enhance the model's capacity to capture different visual features. Third, while Integrated Gradients provided valuable insights into model interpretation, the choice of baseline images and the complexity of the model could influence the reliability of these interpretations. Future research should investigate additional explainable AI techniques to provide a more comprehensive understanding of how visual features influence memorability predictions.

Finally, future work should explore the broader application of our findings to real-world contexts, such as advertising, education, and media. While our work offers significant potential for optimizing visual content for memorability, ethical considerations must be addressed, particularly in fields where memorability could be exploited to manipulate viewer behavior. Researchers and practitioners should follow ethical guidelines to ensure that AI systems designed to predict memorability are used responsibly.

Our findings have direct applications in AI-driven content generation, advertising, and educational media, where optimizing visual elements for memorability can enhance engagement and information retention. In digital marketing, AI systems could generate ad creatives that maximize recall, while in education, instructors could design learning materials that leverage memorability cues for improved knowledge retention. In medical imaging, this approach could be used to highlight diagnostically important regions by optimizing the memorability of critical visual patterns.

By integrating cognitive science principles with AI-based feature extraction, this study contributes to both theoretical understanding and practical applications in AI-driven memorability modeling. These findings pave the way for novel computational tools that evaluate and enhance visual content based on memorability, ultimately improving user engagement, information retention, and human-computer interaction strategies.

\section*{Data Availability}
The datasets used in this study are publicly available. MemCat can be downloaded from the \href{http://gestaltrevision.be/projects/memcat/}{MemCat dataset website}, and ImageNet is available at the \href{https://www.image-net.org/download.php}{ImageNet download page}.

% The datasets generated during and/or analysed during the current study are available in the [NAME] repository, [PERSISTENT WEB LINK TO DATASETS]./
% --- References Section ---
\bibliography{Manuscript.bib}

% For data citations of datasets uploaded to e.g. \emph{figshare}, please use the \verb|howpublished| option in the bib entry to specify the platform and the link, as in the \verb|Hao:gidmaps:2014| example in the sample bibliography file.

% \section*{Acknowledgements (not compulsory)}
\section*{Acknowledgments}
This work was supported by the Natural Sciences and Engineering Research Council of Canada (NSERC) Discovery Grant to Y.M. and a Vector Institute Postdoctoral Fellowship to E.B.

% Acknowledgements should be brief, and should not include thanks to anonymous referees and editors, or effusive comments. Grant or contribution numbers may be acknowledged.

\section*{Author contributions statement}
E.B. and Y.M. contributed to the conceptualization and design of the study. E.B. was primarily responsible for model development, data analysis, and drafting the manuscript. Y.M. provided supervision, methodological guidance, and critical revision of the manuscript. Both authors reviewed and approved the final version of the manuscript.
% Must include all authors, identified by initials, for example:
% A.A. conceived the experiment(s),  A.A. and B.A. conducted the experiment(s), C.A. and D.A. analysed the results.  All authors reviewed the manuscript. 

\section*{Competing interests} The authors declare that they have no conflicts of interest related to this work.
% \section*{Additional information}
% To include, in this order: \textbf{Accession codes} (where applicable); \textbf{Competing interests} (mandatory statement). 
% \textbf{Accession codes:} Not applicable.\\[5pt]

% The corresponding author is responsible for submitting a \href{http://www.nature.com/srep/policies/index.html#competing}{competing interests statement} on behalf of all authors of the paper. This statement must be included in the submitted article file.

\end{document}